\documentclass[pdflatex,sn-mathphys-num]{sn-jnl}

\usepackage{graphicx}%
\usepackage{multirow}%
\usepackage{amsmath,amssymb,amsfonts}%
\usepackage{amsthm}%
\usepackage{mathrsfs}%
\usepackage[title]{appendix}%
\usepackage{xcolor}%
\usepackage{textcomp}%
\usepackage{manyfoot}%
\usepackage{booktabs}%
\usepackage{algorithm}%
\usepackage{algorithmicx}%
\usepackage{algpseudocode}%
\usepackage{listings}%
\usepackage{graphicx}%
\usepackage{longtable}%
\usepackage{array}%
\usepackage{geometry}%

\begin{document}

\title[Article Title]{Principles and Components of Federated Learning Architectures}



\author*[1]{\fnm{MD Abdullah Al} \sur{Nasim}}\email{nasim.abdullah@ieee.org}

\author[2]{\fnm{Fatema Tuj Johura}\sur{Soshi}}\email{fatemasoshi@gmail.com}

\author[3]{\fnm{Parag} \sur{Biswas}}\email{text2parag@gmail.com}

\author[4]{\fnm{ A.S.M Anas} \sur{Ferdous}}\email{anasferdous001@gmail.com}

\author[5]{\fnm{Abdur} 
\sur{Rashid}}\email{rabdurrashid091@gmail.com}

\author[6]{\fnm{Angona} \sur{Biswas}}\email{angonabiswas28@gmail.com}

\author[7]{\fnm{Kishor} \sur{Datta Gupta}}\email{kgupta@cau.edu}



\affil[1,6]{\orgdiv{Research and Development Department}, \orgname{Pioneer Alpha},  \orgaddress{{\city{Dhaka}, \country{Bangladesh}}}}

\affil[2]{\orgdiv{Msc in Data Science and Analytics}, \orgname{University of Hertfordshire},  \orgaddress{{\city{Hatfield}, \country{UK}}}}

\affil[4]{\orgdiv{Department of Biomedical Engineering}, \orgname{Bangladesh University of Engineering and Technology}, \orgaddress{ \city{Dhaka},  \country{Bangladesh}}}

\affil[3, 5]{\orgdiv{MSEM Department}, \orgname{Westcliff university},  \orgaddress{{\city{California}, \country{United States}}}}

\affil[7]{\orgdiv{Department of Computer and Information Science}, \orgname{Clark Atlanta University}, {\city{Georgia}, \country{USA}}}


\abstract{Federated Learning (FL) is a machine learning framework where multiple clients, from mobiles to enterprises, collaboratively construct a model under the orchestration of a central server but still retain the decentralized nature of the training data. This decentralized training of models offers numerous advantages, including cost savings, enhanced privacy, improved security, and compliance with legal requirements. However, for all its apparent advantages, FL is not immune to the limitations of conventional machine learning methodologies. This article provides an elaborate explanation of the inherent concepts and features found within federated learning architecture, addressing five key domains: system heterogeneity, data partitioning, machine learning models, communication protocols, and privacy techniques. This article also highlights the limitations in this domain and proposes avenues for future work. Besides, we provide a set of architectural patterns for federated learning systems, which are derived from the systematic survey of the literature. The main elements of FL, the fundamentals of Federated Learning, and a few architectural specifics will all be better understood with the aid of this research.}

\keywords{Federated Learning, Federated Learning Architectures, Machine Learning, Vertical Federated Learning, Horizontal Federated Learning, FEDF framework}

\maketitle

\section{Introduction}\label{sec1}

The fields of artificial intelligence (AI) and machine learning (ML) have gained tremendous growth in the past few years, especially due to advancements in autonomous driving technologies by players like Tesla and Waymo \cite{badue2021self}. The availability of large datasets and high-powered computing resources has enabled the use of machine learning techniques in multiple domains, such as banking, healthcare \cite{biswas2021brain}, \cite{biswas2023hybrid}, \cite{biswas2023active}, \cite{biswas2023generative}, transportation \cite{idris2024cognitive}, customer service \cite{prabadevi2023customer}, e-commerce, and smart home technologies \cite{yang2023robust}. Considering extensive use of machine learning techniques, their security and privacy is essential to protect. The majority of machine learning frameworks build the model by collecting data from different devices or organizations on a centralized server or cloud-based system. This is a significant limitation, especially in the case of security threats in the training dataset because of the sensitive nature of the data it contains. Many hospitals can combine their data to create a shared machine-learning model for detecting breast cancer \cite{khalid2023breast}, \cite{zhao2023clinical} using MRI images. On the other hand, patient data sharing with a centralized server risks the exposure of confidential data, and many negative consequences would then follow. In some situations, Federated Learning would be a preferable option. Federated Learning is a collaborative type of learning in which devices or organizations share and aggregate the model parameters of their local models, instead of local data sharing \cite{savazzi2021opportunities}.

\begin{figure}
\centering
\includegraphics[height=7 cm]{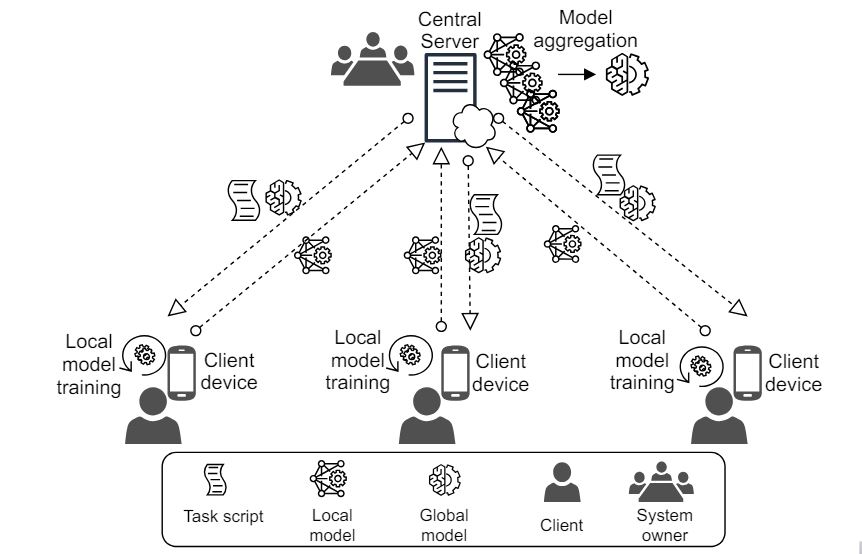}
\caption{General Outlook of Federated Learning
 \cite{lo2022architectural}}
\label{1}
\end{figure}

FL has profoundly impacted machine learning, particularly in terms of data security and privacy management \cite{mothukuri2021survey}. Clients can range from mobile devices to entire businesses. By ensuring that the training data is decentralized, this strategy helps to reduce the hazards that come with sharing data, which is a feature of typical centralized machine learning techniques. FL has especially great promise in industries like finance and healthcare, where data protection and sensitivity are vital. The overview of FL is shown in Fig. \ref{1}. An FL system has three stakeholders: (1) the system owner or learning coordinator; (2) the contributor client, which includes local model trainers and data contributors; and (3) the user client, which is the model user \cite{lo2022architectural}. Keep in mind that a user client can also be a contributor client. System nodes, or hardware components, come in two varieties: (1) central servers and (2) client devices.

Google originally presented the concept of FL in 2016 when they implemented it in the Google Keyboard, enabling several Android phones to learn together. Because FL may be implemented on any edge device, it has the potential to completely transform a number of important industries, including finance, healthcare, transportation, and smart homes \cite{omoniwa2018fog}. The most well-known instance was when scientists and doctors from various countries worked together to create an AI pandemic engine for COVID-19 diagnosis \cite{imran2020ai4covid} using chest scans. Transportation networks present yet another intriguing use case: teaching cars to drive themselves and create city routes. In a similar vein, FL frameworks enable edge devices in various homes to cooperatively learn on context-aware policies for smart-home applications \cite{rentero2022using}.

Therefore, when privacy, bandwidth, and regulatory compliance are important factors, FL provides a number of advantages to centralized machine learning.  FL makes it possible to train models directly on local devices in industries like healthcare and finance, guaranteeing that private information stays on-site and reducing the legal issues that come with data centralization.  By enabling local model training, FL lessens the need to send massive amounts of raw data in the context of the Internet of Things (IoT) and 5G applications, saving bandwidth and improving efficiency. Furthermore, by enabling businesses to create reliable models without moving personal data to central servers, FL complies with strict data protection laws like GDPR and HIPAA.  Google's Gboard is a real-world example of how FL is used; it uses FL to enhance typing predictions on users' devices without sending private text data to central servers \cite{chalamala2022federated}

FL functions based on two fundamental concepts: model transmission and local computing \cite{bouzinis2021wireless}. Clients use their data to do local training; they only provide the trained model parameters to the central server, which combines them to update the global model \cite{agrawal2022federated}, \cite{nguyen2021federated}. Until a workable model is produced, this iterative process is continued. Although FL dramatically lowers some operating expenses and systemic privacy issues, it also presents some special difficulties, such as the requirement for complex coordination and communication methods and vulnerability to a variety of attacks \cite{bouacida2021vulnerabilities}.

\begin{figure}
\centering
\includegraphics[height= 6.5 cm]{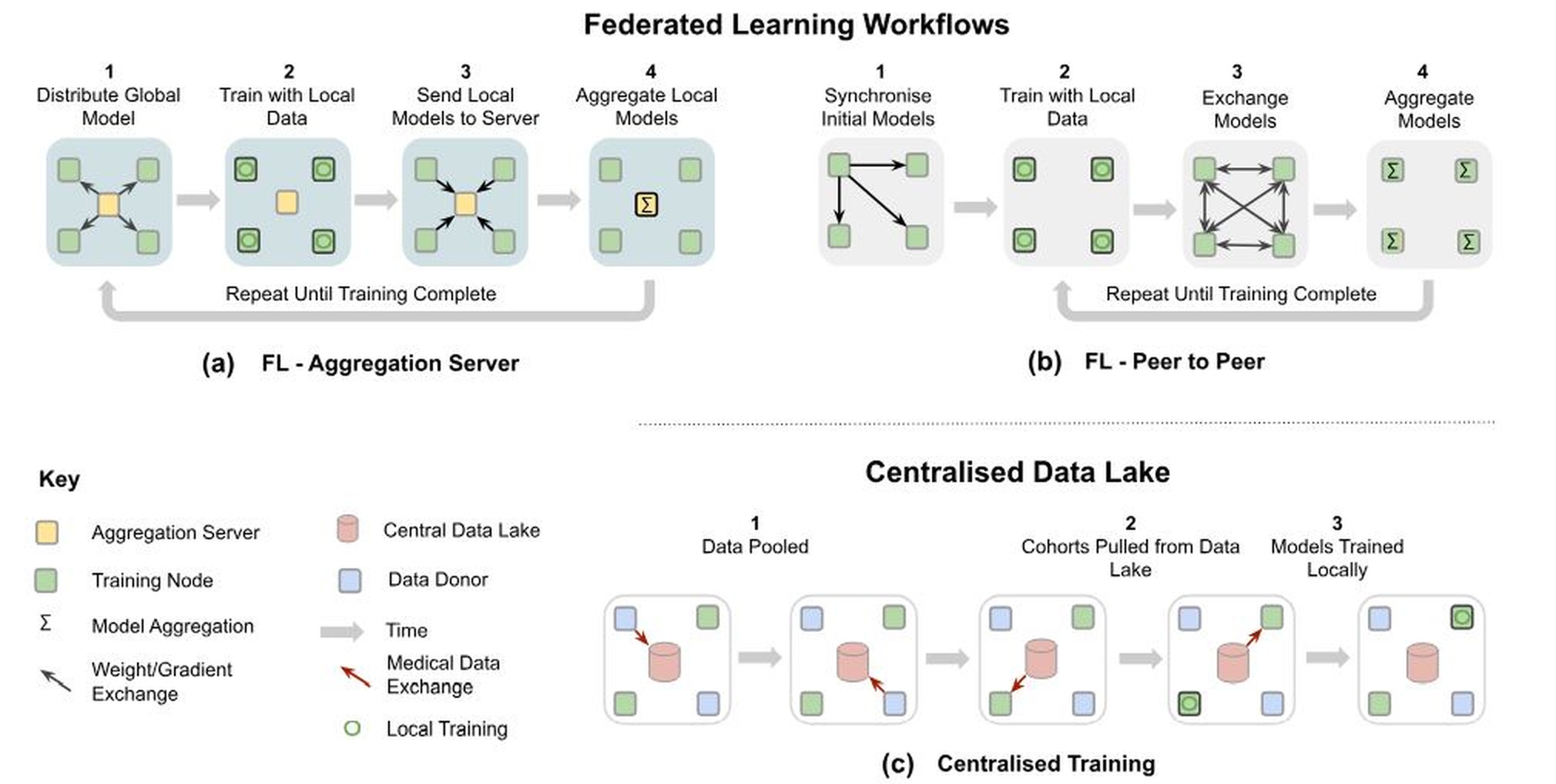}
\caption{The standard FL workflow involves a federation of training nodes that receive a global model, periodically send partially trained models to a central server for aggregation, and continue training on a consensus model provided by the server. We call this process the FL aggregation server (a). (b) FL Peer-to-Peer: An alternative FL formulation where each training node performs its own aggregation and shares its partially learned model with some or all of its peers. A basic non-FL training approach, known as "centralized training" (c), involves data collection sites providing data to a central data lake, from which they retrieve data for independent local training \cite{rieke2020future}}. 
\label{2}
\end{figure}

The authors of the research paper \cite{rieke2020future} illustrate (FL) operations and highlight the distinctions between learning on a centralized data lake and on a workstation, which is shown in Fig. \ref{2}. The study focuses on how FL may be used to incorporate data-driven machine learning (ML), particularly deep learning (DL), into medical practice. The paper examines the difficulty of exploiting massive amounts of medical data due to data silos and privacy concerns, and FL is proposed as a potential solution. FL addresses privacy concerns while also enabling the creation of trustworthy and accurate machine learning (ML) models for the healthcare business without the need for centralized data collection for collaborative model training. The primary contribution of this study is to investigate FL as a viable solution to overcome data silos and privacy concerns in the application of ML for digital health.

\begin{figure}
\centering
\includegraphics[height=4.8cm]{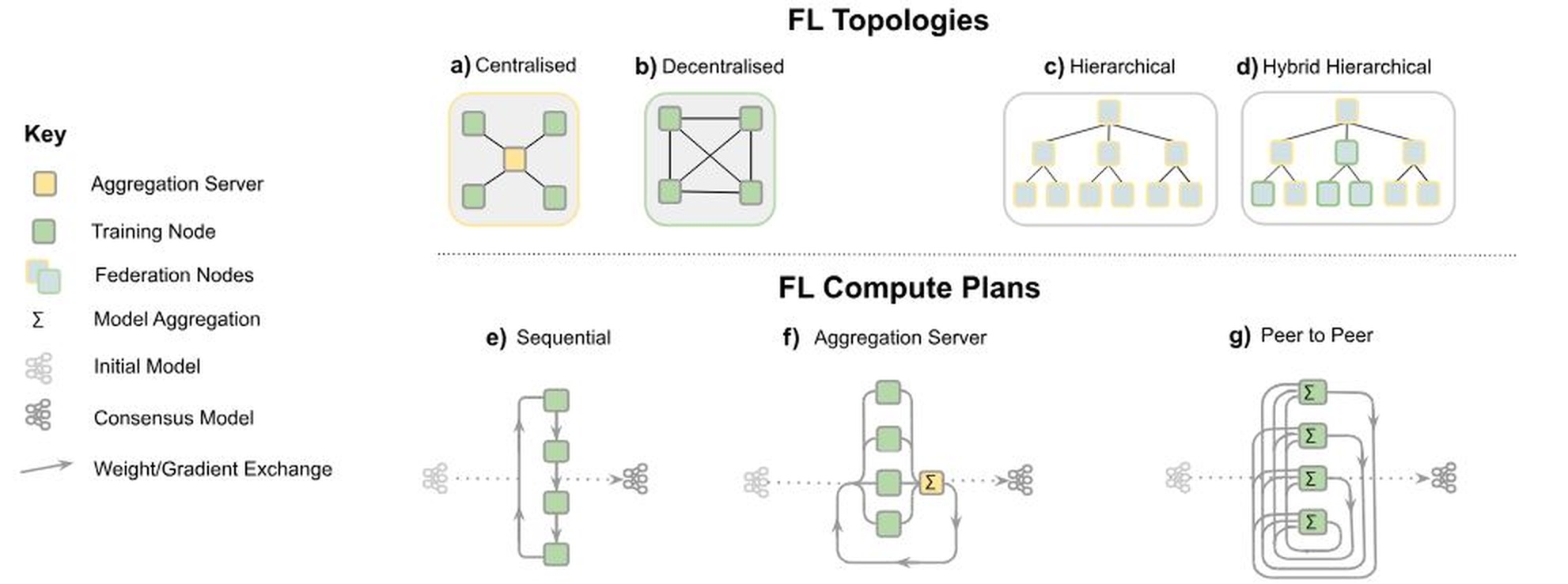}
\caption{Overview of different FL theme options. FL topology: communication architecture of federation. (a) Centralized:  models are collected, aggregated, and distributed among training nodes (hub and spokes) by an aggregation server that also manages the training iterations. (b) Distributed: aggregation happens simultaneously at each training node connected to one or more peers. (c) Hierarchical: Peer-to-peer federations and aggregation server federations can be combined to create various sub-federations forming a federated network (d). FL computation plan: Passing the model through multiple partners. Cycles of transfer learning and sequential training. (f) Peer-to-peer, (g) aggregation server \cite{rieke2020future}}. 
\label{3}
\end{figure}

Figure \ref{3} illustrates the various topologies and computation strategies that may be used to achieve an FL process. Peer-to-peer is the most preferred technique for healthcare applications, followed by an aggregate server. Because FL participants only get model parameters that are averaged among a group of participants and never have direct access to data from other institutions, FL always implicitly ensures a degree of anonymity.

\begin{figure}
\centering
\includegraphics[height= 7 cm]{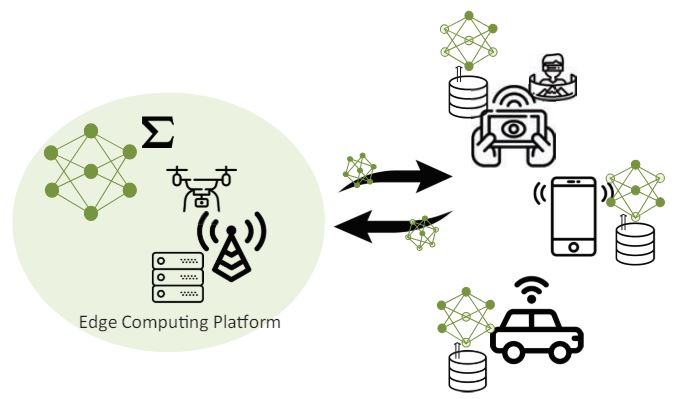}
\caption{An example of federated learning running on caches and edge computing, where the aggregator can be an edge computing platform on an edge network (such as a wireless base station or an unmanned aerial vehicle) and the local learner can be an edge user (an autonomous vehicle in an autonomous vehicle network, or an augmented/virtual platform for a user reality headset) \cite{niknam2020federated}}. 
\label{4}
\end{figure}

The focus of the research paper\cite{niknam2020federated}  is on FL as a distributed and privacy-preserving approach to solving wireless communication problems, especially in the context of fifth-generation (5G) networks. The study highlights the shortcomings of conventional, centralized machine learning (ML) techniques in wireless applications because of serious communication costs and worries about data privacy. Federated learning is offered as a possible solution to improve various wireless communication applications \cite{gao2023federated}, \cite{gebremariam2023blockchain}, and avoid these problems by enabling local model training without centralizing data. The study illustrates the appropriateness of FL for a range of use cases by discussing numerous possible implementations of the technology within 5G networks \cite{kazmi2024security}. FL, which uses locally trained models instead of gaining direct access to user data, appears to be a perfect fit for proactive caching in wireless networks, specifically for content popularity prediction, as shown in Fig. \ref{4}

The aim of this study is to provide a comprehensive overview of the basic ideas and elements of FL architecture. We explore the key components that characterize FL systems, such as data partitioning tactics, privacy-preserving methods, the kinds of machine learning models used, communication protocols, and the management of heterogeneity within the system. Moreover, we recognize and talk about architectural patterns that provide reusable fixes for typical design issues that arise during FL system development. This review aims to improve knowledge and application of FL systems by summarizing ongoing research and suggesting future research avenues. In order to advance the discipline and meet the inherent issues of designing safe, reliable, and efficient FL systems, we hope to be of great assistance to researchers and practitioners.

\section{Basic Principles of Federated Learning}

Federated Learning is a different way to use computers to learn from data. Instead of sending all their information to one place, devices like phones or computers learn on their own. They only send small updates to a main server. This keeps people's information private and helps save money. Below are the main ideas about Federated Learning.

\subsection{Localization of Data and Decentralization}


\subsubsection{Local Data Utilization} FL prevents sensitive data from being sent to a central server by keeping raw data on the client's device or local system.  Every client uses its own data to independently carry out local calculations and model training \cite{chahoud2023feasibility}.


\subsubsection{Decentralized Data Storage} FL greatly improves security and privacy by preserving decentralized data storage.  This strategy lowers the possibility of data breaches and guarantees adherence to stringent laws in sectors like healthcare and finance.   FL reduces these risks in contrast to centralized ML, where data aggregation raises exposure to cyberthreats \cite{qu2020decentralized}.

\subsection{Collaborative Model Training}

\subsubsection{Federated Training Process} 

FL keeps the data private while allowing for teamwork in training models on different devices or by different organizations. Each participant uses their own data to calculate and send updates to a main server. To protect your data, only encrypted updates to the model are sent, not the raw data itself. \cite{han2023practical}.

\subsubsection{Model Aggregation} 

The central server improves a global model by collecting, processing, and combining updates from multiple clients. This process continues until the model is stable. Using techniques like Federated Averaging (FedAvg), which combines knowledge from different datasets, helps make the model more effective.

\subsection{Privacy Preservation}

\subsubsection{No Raw Data Sharing} Federated learning helps solve privacy problems better than centralized machine learning (ML) by letting clients share only model updates, like changes to parameters or gradients, instead of raw data. This greatly lowers the chance of exposing sensitive information during training. It also allows clients to keep control of their data, addressing important issues related to data ownership and privacy laws. By focusing on combining model updates, federated learning can reduce biases that may come from merging sensitive datasets. The decentralized structure of federated learning not only boosts privacy but also builds trust among participants, encouraging more people to take part in collaborative training. This makes federated learning a useful way to create strong and privacy-friendly machine learning systems.

\subsection{Strategies for Improving Privacy} To further protect the confidentiality of the shared updates, strategies such as homomorphic encryption, secure multi-party computation, and differential privacy can be used.

\subsection{Communication Efficiency}

\subsubsection{Optimized Communication Protocols} 

In FL, efficient communication protocols are crucial for reducing latency and bandwidth usage.  Protocols that are optimized minimize synchronization delays and provide seamless communication between clients and the central server \cite{mills2022client}.

\subsubsection{Model Compression and Update Sparsification:} 

Due to the possibility of high communication costs from frequent model updates, FL uses compression techniques like:
\begin{itemize}
    \item \textbf{Quantization: } Reduces the precision of model parameters to lower data transmission costs
    \item \textbf{Update sparsification: } Reduces the amount of data that is transmitted by choosing only the most important parameter changes. 

\end{itemize}

\subsection{Handling Data and System Heterogeneity}

\subsubsection{Managing Diverse Data Distributions:} 
Client data in federated learning applications often varies significantly and is not uniform. This can harm the performance of the overall model because different clients may have very different data patterns. FL algorithms must be adjusted to handle these variations effectively.

\subsubsection{Addressing Client and System Variability} 
Many client devices have different processing abilities, energy levels, and network conditions that must be supported by federated learning systems. Techniques like asynchronous aggregation and customized training schedules help make it easier for devices with limited resources to participate effectively.

\subsection{Security and Robustness}

\subsubsection{Defense Against Model Poisoning Attacks} 

Adversarial threats such as model poisoning, in which malevolent clients introduce modified updates to taint the global model, must be thwarted by FL. Such risks are lessened by methods like anomaly detection and Byzantine-resilient aggregation.

\subsubsection{Fault Tolerance and Resilience} 

Unreliable network connections and client dropouts must be supported by FL systems.  Techniques like redundant calculations and backup models guarantee that the system is robust even if certain clients do not show up for training sessions.

\subsection{Scalability Considerations}

FL needs to be built with efficiency in mind, supporting millions of clients.  Important scalability enhancements consist of:
\begin{itemize}
    \item \textbf{Hierarchical FL: } It distributes the workload by introducing intermediary aggregation nodes between the clients and the central server.

    \item \textbf{Edge computing integration: } Permits edge servers to submit updates to the cloud after completing partial aggregation.

\end{itemize}

\section{Explainable AI (XAI) and Zero Trust Architecture (ZTA)}
Integrating Explainable AI (XAI) into Zero Trust Architecture (ZTA) is essential for maintaining security, accountability, and trust in federated learning (FL) systems. ZTA follows the principle of "never trust, always verify," which requires continuous re-evaluation of authentication and access decisions. XAI provides transparency by explaining why requests are approved or denied and identifying the causes of unusual behavior, ensuring compliance with security policies.

For example, if a hospital node is flagged for suspicious data, XAI can reveal specific features, like abnormal timestamps or value ranges, helping auditors understand the alert. In finance, XAI enhances transaction monitoring by highlighting unusual transfer amounts or geolocation patterns, helping to detect fraud. \cite{guembe2022explainable}.

\subsection{XAI for Access Control and Anomaly Detection}
In a zero-trust, federated environment, interpretability is crucial. When a user or device requests access, an explainable AI (XAI) system clarifies the decision-making process. It can reveal whether access was denied due to the device's security posture or if the user's authentication history warranted extra scrutiny. By presenting this information in clear terms, XAI turns complex decisions into actionable insights.

For anomaly detection, XAI models analyze large volumes of network data to identify deviations from established baselines, providing contextual explanations for these anomalies. This allows security analysts to quickly understand issues, whether it’s unusual traffic to an unknown port or unexpected device communication.

Additionally, these explanations are logged, creating a verifiable record that proves compliance with Zero Trust Architecture (ZTA) principles. In 5G network security, for example, combining federated learning (FL) with XAI enables operators to learn normal traffic behaviors in real time and quickly identify and explain any deviations, catching intrusions early. \cite{gummadi2024xai}.

\subsection{Federated Learning’s Role in Zero Trust Architecture (ZTA)}
Federated Learning supports Zero Trust principles by decentralizing model training and reducing the reliance on a central data store. Instead of collecting sensitive information in one place, it lets each node like a hospital database or edge device train locally and only send updates about the model. This method improves data confidentiality, lowers the risk of attacks, and establishes strict controls over sensitive information.

Additionally, real-time behavior analysis helps adjust trust levels in the system. Nodes that follow security rules earn high trust scores, while those that act suspiciously lose trust. Overall, the ongoing learning process of Federated Learning, together with Explainable Artificial Intelligence (XAI), aligns with the focus of Zero Trust Architecture (ZTA) on continuous verification and limited access \cite{hussain2024federated}. 

\section{Evaluation Of The Performance Of Federated Learning Algorithms}

The main focus of the study \cite{nilsson2018performance} is the evaluation and comparison of several federated learning algorithms, focusing on their performance on independent identically distributed (i.i.d.) and non-i.i.d datasets. Federated learning is an important distributed machine learning technique that combines locally trained models from data-generating clients, such as connected cars and smartphones, to train a global model. Federated Averaging (FedAvg), Federated Stochastic Variance Reduced Gradient (FSVRG), and CO-OP are the three federated learning techniques compared in this article. The MNIST dataset is used to thoroughly compare the performance of these techniques. Among the tested federated learning algorithms, FedAvg proves to be the most successful, especially when dealing with i.i.d data. In this section, we will discuss these three algorithms taken from the study \cite{nilsson2018performance}.

\begin{figure}
\centering
\includegraphics[height= 5 cm]{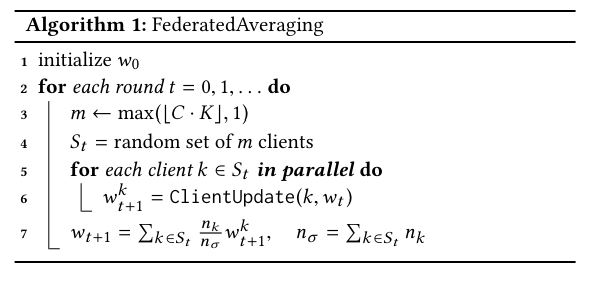}
\label{5}
\end{figure}

FedAvg employs a central server to assist training by hosting the shared global model \(w_t \), where \(t \) specifies the communication round. Nonetheless, true optimization is performed locally on clients using technologies such as Stochastic Gradient Descent (SGD). FedAvg's five hyperparameters are the proportion of customers to train (C), the local mini-batch size (B), the number of local epochs (E), a learning rate \(eta\), and optionally a learning rate decay \(lambda\). SGD training typically uses the parameters \(B \), \(E \), \(\eta \), and \(\lambda \). However, in this scenario, \(E \) indicates the total number of iterations over the local data before an update to the global model. 
 The number of local training instances determines the weighting system, as stated in Algorithm \ref{5} on line 7.

\begin{figure}
\centering
\includegraphics[height= 7 cm]{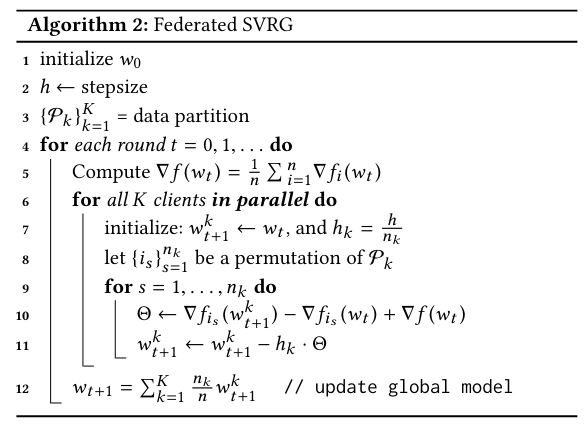}
\label{6}
\end{figure}

FSVRG works by executing multiple distributed stochastic modifications on each client following an expensive central full gradient calculation. To obtain a stochastic update, one update is performed iteratively for each data point, using a random permutation of the local data. A basic FSVRG only has one hyperparameter, the stepsize \(h\). Algorithm \ref{6} thoroughly explains FSVRG, which involves a single iteration as follows: To calculate a total gradient, all clients acquire the most recent version of the model and calculate loss gradients in connection to their local data. Clients then submit their gradients, which the server aggregates to generate the entire gradient \(\nabla f (w_t) \).

Unlike FedAvg and FSVRG, which rely on coordinated model updates, CO-OP \cite{wang2017co} proposes an asynchronous approach. This technique instantaneously combines any incoming client model with the global model. The global model has age \(a \), and each client \(k \) has an age \(a_k \) associated with it. When merging models, the equation for the age difference  \( a - a_k \)) is used to compute a weight. The justification for this is that in an asynchronous structure, some clients will train on out-of-date models while others will train on newer models.

\begin{figure}
\centering
\includegraphics[height= 9 cm]{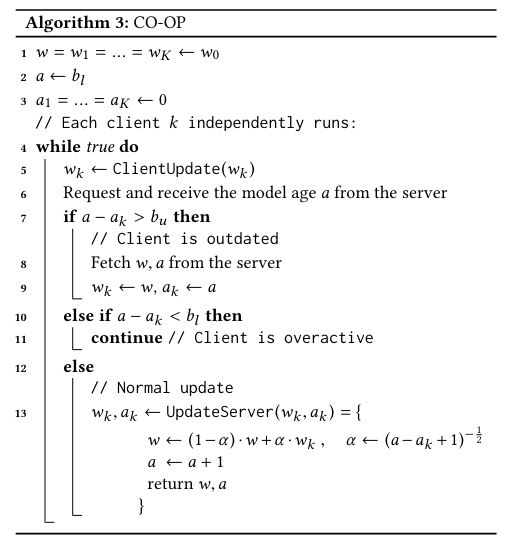}
\label{7}
\end{figure}

Additionally, CO-OP gets all of its hyperparameters from the optimization technique that underpins it, such as SGD.
The following is the training protocol: Using its own training set of data, each client runs an optimization process over E rounds before asking the server for the global model age as of right now. At this point, the client determines whether the age gap satisfies the requirements. In the event that the local model is out of date, the client makes amends with the global model and restarts. In the event that the client exhibits excessive activity, training just continues. If not, the local model is uploaded to the server in order to be combined. We see the CO-OP pseudocode in Algorithm \ref{7}.

\section{Federated Learning Architectures}

FL is an approach that allows machine learning models to be trained in a distributed manner using remotely hosted datasets without polluting the data through aggregation.
 \cite{antunes2022federated}. FL is a viable way to improve ML-based systems, increase alignment with regulatory standards, and improve data sovereignty and trust. Several challenges remain, including data heterogeneity, privacy risks, communication efficiency, and secure model aggregation, which must be addressed before FL achieves widespread adoption. Both federated learning and neural architecture search face many unsolved challenges. However, the search for optimal neural designs in the context of federated learning is particularly challenging \cite{zhu2021federated}. This work provides background on Federated Learning and Neural Architecture Search (NAS), with a particular focus on the recently developed area of Federated Neural Architecture Search (FNAS). Systems are categorized into offline and online approaches, and single- and multi-objective NAS methods are discussed. The study classifies federated learning systems, draws attention to the difficulties and limitations of online FNAS, looks at ways to balance various goals including precision and communication expenses, and concludes by summarizing the primary issues still facing FNAS.

 FL's application base has grown over time to include a wide range of fields, including banking, healthcare, and other industries. The creation of novel algorithms, privacy-preserving methods \cite{yin2021comprehensive}, and designs to manage the inherent heterogeneity in federated environments are important developments.

The study \cite{aledhari2020federated} provides a comprehensive examination of FL, emphasizing privacy-preserving solutions and focusing on enabling technologies, protocols, practical implementations, and use cases across various sectors. It intends to help data scientists create more effective privacy-preserving solutions by offering a complete review of relevant FL protocols, platforms, and applications. It also discusses the primary advantages and disadvantages of FL, as well as detailed use cases that demonstrate how successfully it can be utilized in various industries.

\begin{figure}
\centering
\includegraphics[height= 5 cm]{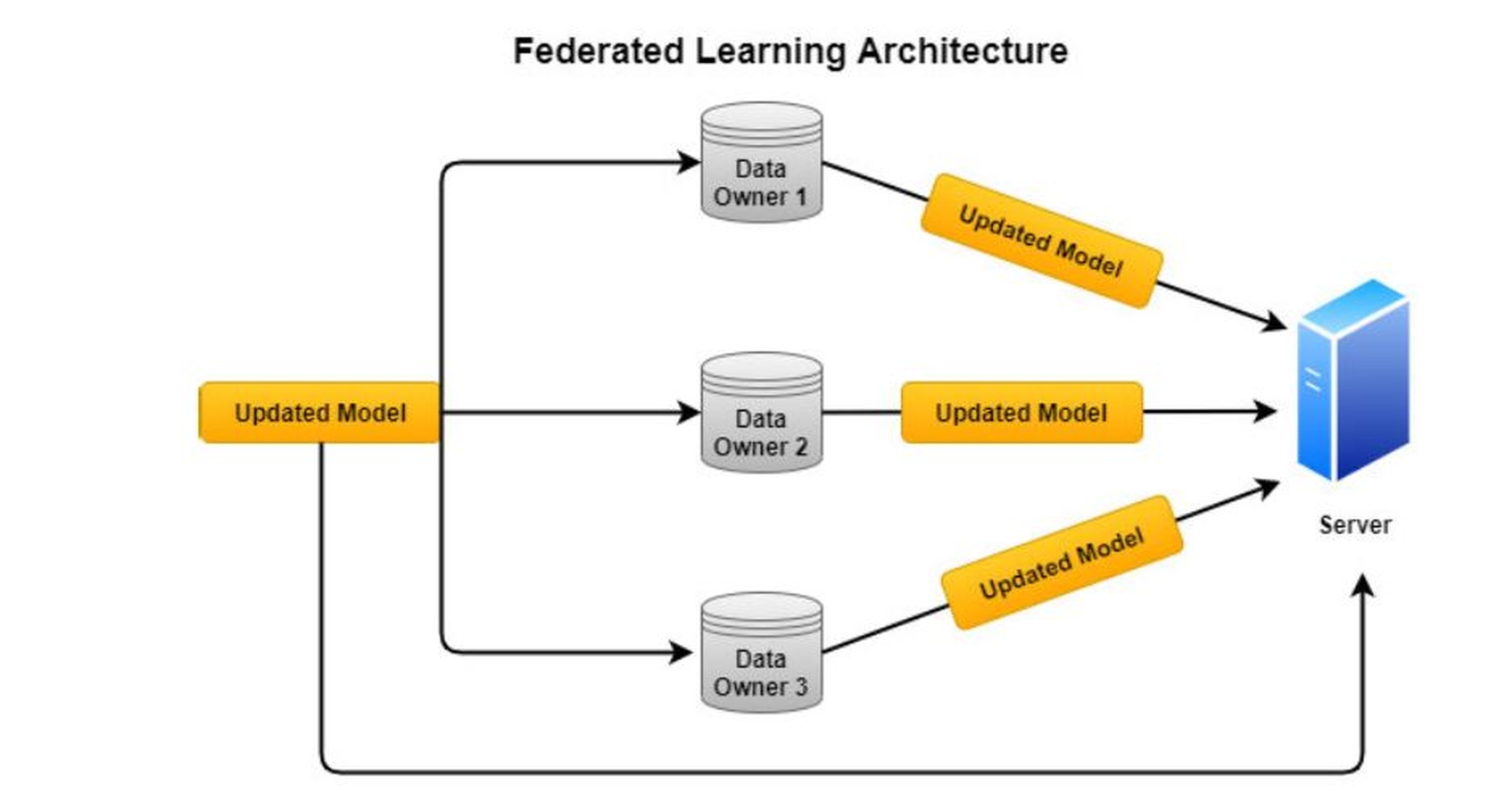}
\caption{Universal architecture for federated learning \cite{aledhari2020federated}.}
\label{8}
\end{figure}

The revised models are returned to the principal server for aggregation. The devices receive a single, aggregated model based on distributed computing principles \cite{ben2021deep}. This enables us to monitor and disperse each model among several devices. FL's technique is particularly advantageous for using affordable machine learning models on devices such as sensors and mobile phones \cite{doku2019towards}. Figure \ref{8} exhibits FL's general architecture. 

\begin{figure}
\centering
\includegraphics[height= 5 cm]{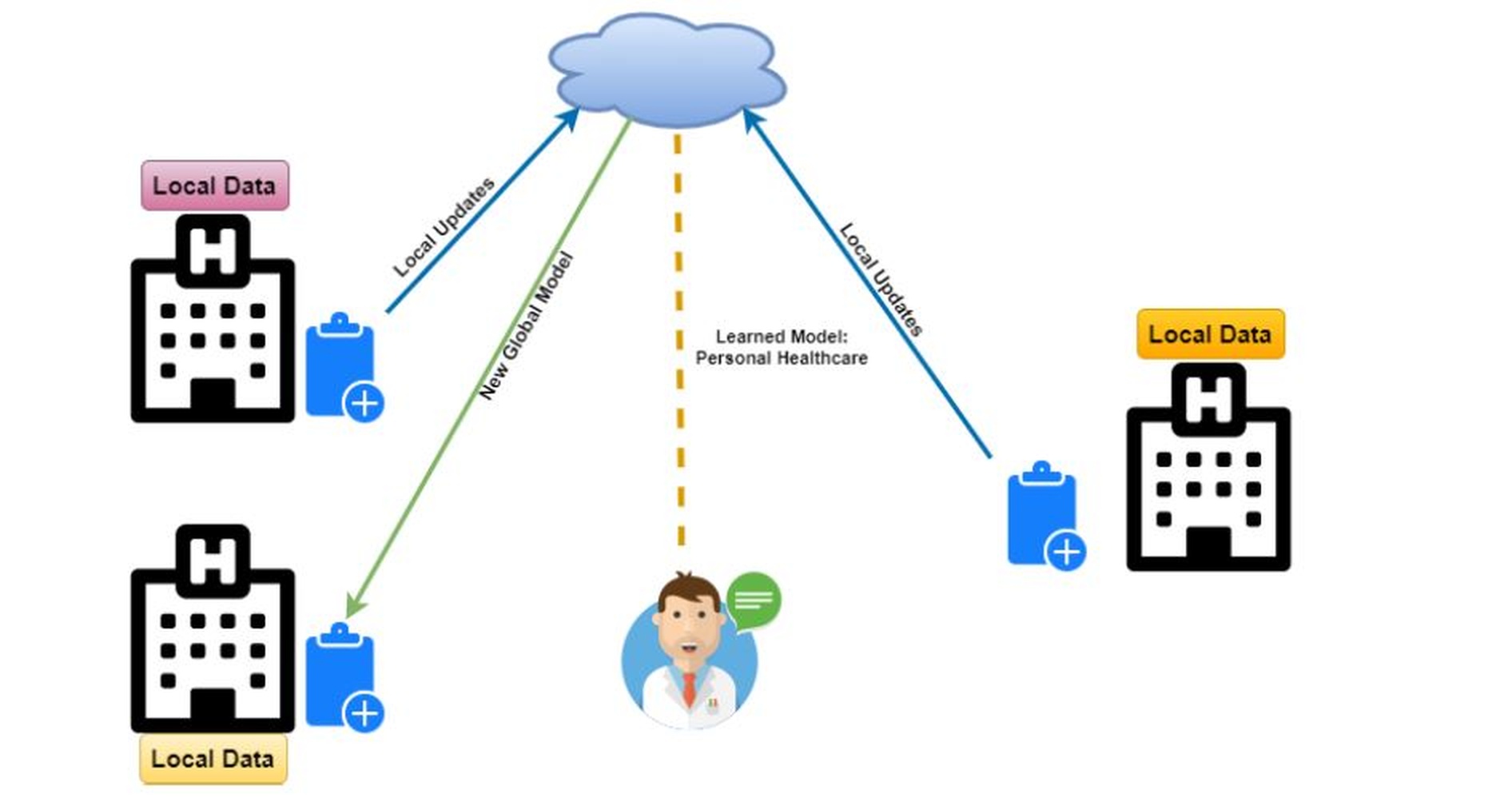}
\caption{Utilizing Federated Learning Architecture in a Healthcare Environment \cite{aledhari2020federated}.}
\label{9}
\end{figure}

There is a wealth of research on the usage of FL. One of its unique use cases is the healthcare industry \cite{stoian2008current}, \cite{brisimi2018federated}. Figure \ref{9} illustrates the use of a FL design in a hospital context. Unfortunately, there are still considerable impediments to FL's full integration in other situations, notably with regard to data. 

The study \cite{singh2022federated} examines the field of federated learning, highlighting its potential in a number of industries and its use in mobile devices. It explores the different forms, structures, possibilities, and difficulties associated with federated learning with the goal of creating common practices for broad deployment in dispersed settings, protecting data, and facilitating diverse networks. It seeks to offer a road map for federated learning adoption and use across a range of industries, such as mobile networks, healthcare, and transportation.

\begin{figure}
\centering
\includegraphics[height= 5 cm]{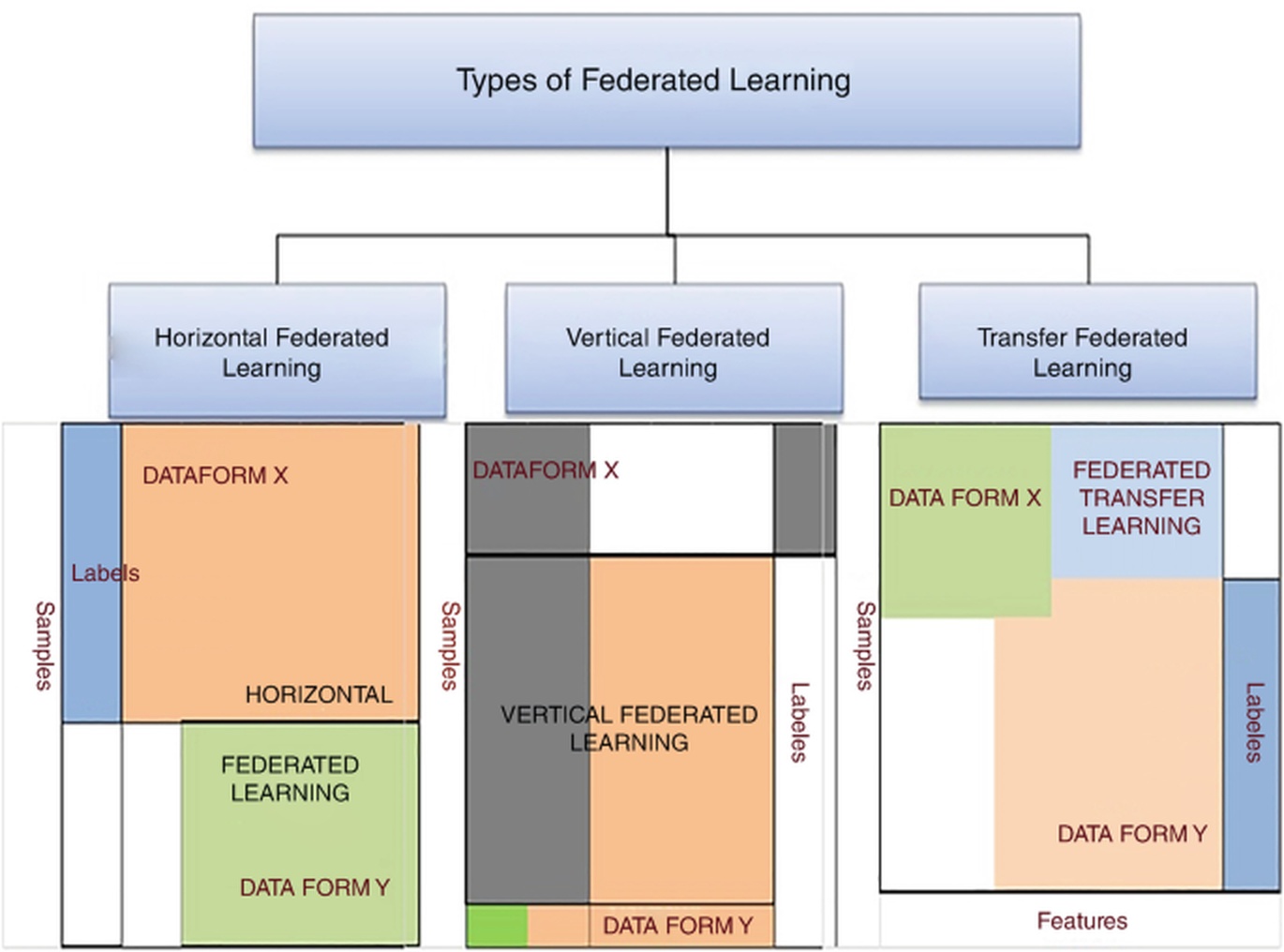}
\caption{Some major types of Federated Learning Architecture are shown by the authors \cite{singh2022federated}.}
\label{10}
\end{figure}

There are numerous platforms and architectures included with FL.
Numerous organizations are currently working to create FL designs in the medical industry \cite{bonawitz2019towards}, \cite{cheng2021secureboost}. Intel and the University of Pennsylvania are two of the top universities. Furthermore, a variety of platforms have been developed for FL, a few of which will be discussed in this part. Table \ref{tab:fl_architectures} summarizes several designs and their focus points. This section goes into further information regarding these architectures.

\begin{longtable}{|>{\centering\arraybackslash}m{2.5 cm}|>{\centering\arraybackslash}m{4 cm}|>{\centering\arraybackslash}m{2.5 cm}|>{\centering\arraybackslash}m{3.5cm}|}
\caption{An overview of architectures, a brief synopsis, and their main focus} \label{tab:fl_architectures} \\
\hline
\textbf{Name of FL Architecture} & \textbf{Short Description} & \textbf{Characteristics} & \textbf{Benefits and Main Focus of Application} \\
\hline
Horizontal Federated Learning (HFL) & Federated learning use an identical feature space but distinct sample spaces. & Each client has data with the same features. & Enhances collaboration among institutions with similar data structures. Focus: Healthcare collaboration between hospitals. \\
\hline
Vertical Federated Learning (VFL) & Federated learning use the same sample region but distinct feature spaces. & Each client has different features for the same samples. & Combines complementary data from different domains. Focus: Cross-sector collaboration, e.g., between banks and insurance companies. \\
\hline
Federated Transfer Learning (FTL) & Combines transfer and federated learning for many instances and feature spaces. & Uses pre-trained models to adapt to new tasks or domains. & Facilitates knowledge transfer across domains. Focus: Cross-domain collaborations to improve models. \\
\hline
Centralized Federated Learning & Coordinates the learning process through a central server. & Simplifies aggregation, potential central bottleneck. & Easy to manage but may suffer from central bottlenecks. Focus: General applications with centralized data control. \\
\hline
Decentralized Federated Learning & No central server; clients communicate and share updates directly with each other. & No central bottleneck, no single point of failure. & Increases robustness and fault tolerance. Focus: Applications needing high robustness. \\
\hline
Hierarchical Federated Learning & Introduces intermediate aggregators between clients and the central server. & Reduces central server load, and enhances scalability. & Leverages edge computing for scalability. Focus: Scalable applications using edge devices. \\
\hline
Asynchronous Federated Learning & Clients send updates to the server asynchronously. & Reduces idle time, and handles stragglers effectively. & Optimizes for environments with latency issues. Focus: Real-time applications with intermittent connectivity. \\
\hline
PERFIT & Federated learning for personalized fitness recommendations. & Customizable to individual fitness data and goals. & Provides personalized health insights. Focus: Fitness and health tracking applications. \\
\hline
MMVLF (Multi-Model Vertical Federated Learning) & Enables the training of multiple models vertically. & Combines various feature sets for comprehensive insights. & Enhances model accuracy and robustness. Focus: Multi-domain data analysis and insights. \\
\hline
FADL (Federated Anomaly Detection Learning) & Federated learning tailored for anomaly detection. & Detects anomalies across distributed datasets. & Improves security and fault detection. Focus: Cybersecurity and fraud detection. \\
\hline
Blockchain-FL & Integrates blockchain with federated learning for secure model updates. & Decentralized ledger for verifiable updates. & Enhances security and transparency. Focus: Secure and transparent data collaboration. \\
\hline
FEDF (Federated Edge-Device Framework) & Framework for federated learning on edge devices. & Optimized for resource-constrained environments. & Empowers edge devices with federated learning capabilities. Focus: IoT and mobile device applications. \\
\hline
\end{longtable}

\begin{figure}
\centering
\includegraphics[height= 6 cm]{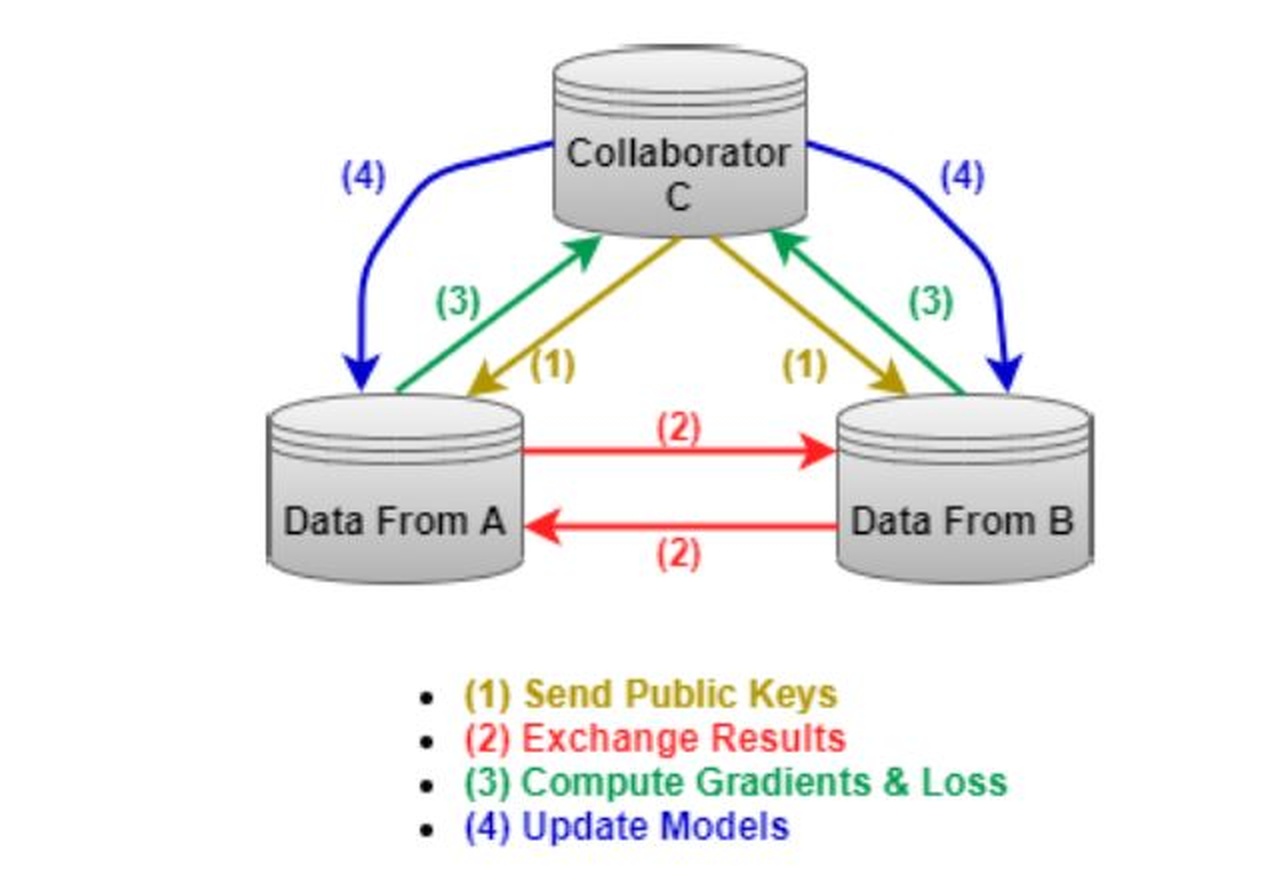}
\caption{The architecture of Vertical Federated Learning \cite{aledhari2020federated}.}
\label{11}
\end{figure}

\begin{figure}
\centering
\includegraphics[height= 6 cm]{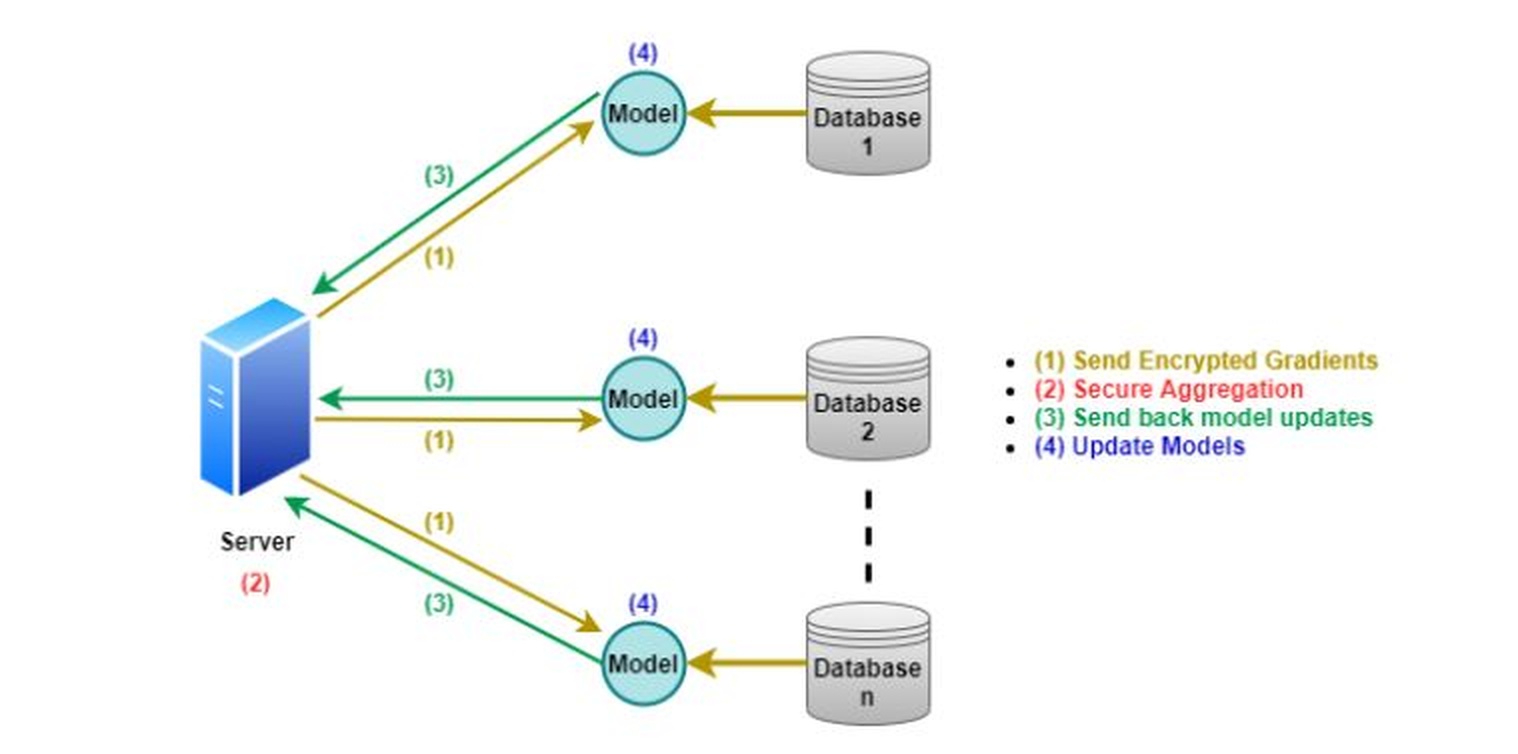}
\caption{The architecture of Horizontal Federated Learning \cite{aledhari2020federated}.}
\label{12}
\end{figure}

FL (also known as sample-based FL) refers to comparable features that differ in terms of data. It's worth noting that ideas for a horizontal FL framework have been made. One example is when Google proposed utilizing a Horizontal FL method to manage Android phone upgrades. Horizontal FL assumes that consumers are trustworthy and that the server is secure. Customer data can only be updated by the central server \cite{bonawitz2019towards}. Horizontal FL's architecture allows x number of analogous structural pieces to learn a model with the support of servers or parameters, as seen in Fig. \ref{12}.

Vertical FL is also known as feature-based FL. Figure \ref{11} depicts the Vertical FL procedure. In this case, data sets may differ in features but have similar sample IDs. What we are doing with Vertical FL is gathering and organizing these different elements. Next, in order to create a model that collectively incorporates data from both entities, we must compute the training loss. Every entity in Vertical FL shares the same identification and status. The Vertical FL system also presumes that its customers are trustworthy when it comes to security. Nonetheless, Vertical FL raises two security-related issues. The Vertical FL architecture consists of two primary components: encrypted model training and encrypted entity alignment \cite{cheng2021secureboost}, \cite{yang2019federated}.

This architecture's independence from other machine-learning techniques is one of its advantages. It's interesting to note that horizontal FL has been applied to medical situations like drug detection. Federated Transfer Learning (FTL) is an additional FL architecture in addition to the Horizontal FL and Vertical Architectures. In \cite{liu2020secure}, FTL was proposed.

\begin{figure}
\centering
\includegraphics[height= 6 cm]{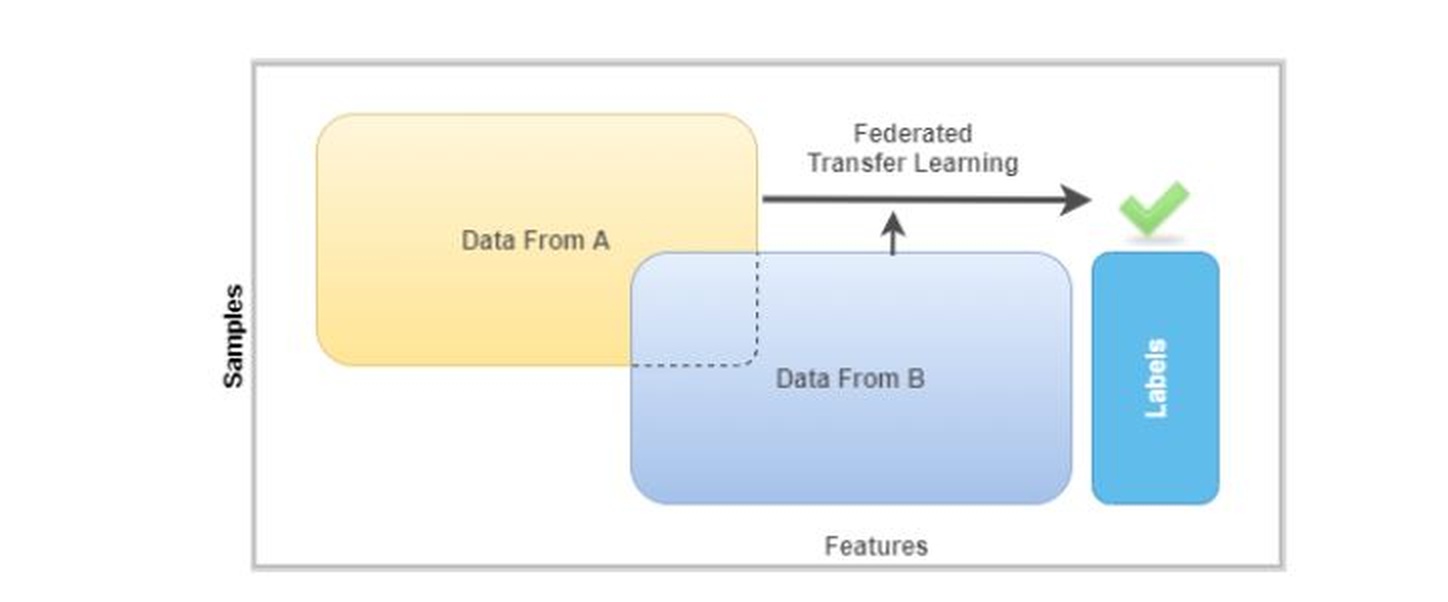}
\caption{The architecture of Federated Transfer Learning \cite{aledhari2020federated}.}
\label{13}
\end{figure}

Figure \ref{13} presents an overview of the FTL procedure. To complete the method, the Guest and Host must first compute and encrypt their findings locally. Gradients and losses are calculated using the data. They then provide Arbiter access to the encrypted values. The Arbiter then provides the Guest and Host with the gradients and loss computations, which they may use to make model adjustments. Until the loss function converges, the FTL framework iterates. Additionally, FTL offers support for both homogenous and heterogeneous training methodologies. Using a variety of sample types, entities assist in training the model or models in the homogeneous method. When entities are heterogeneous, they have identical samples but differing feature spaces. 

\begin{figure}
\centering
\includegraphics[height= 6 cm]{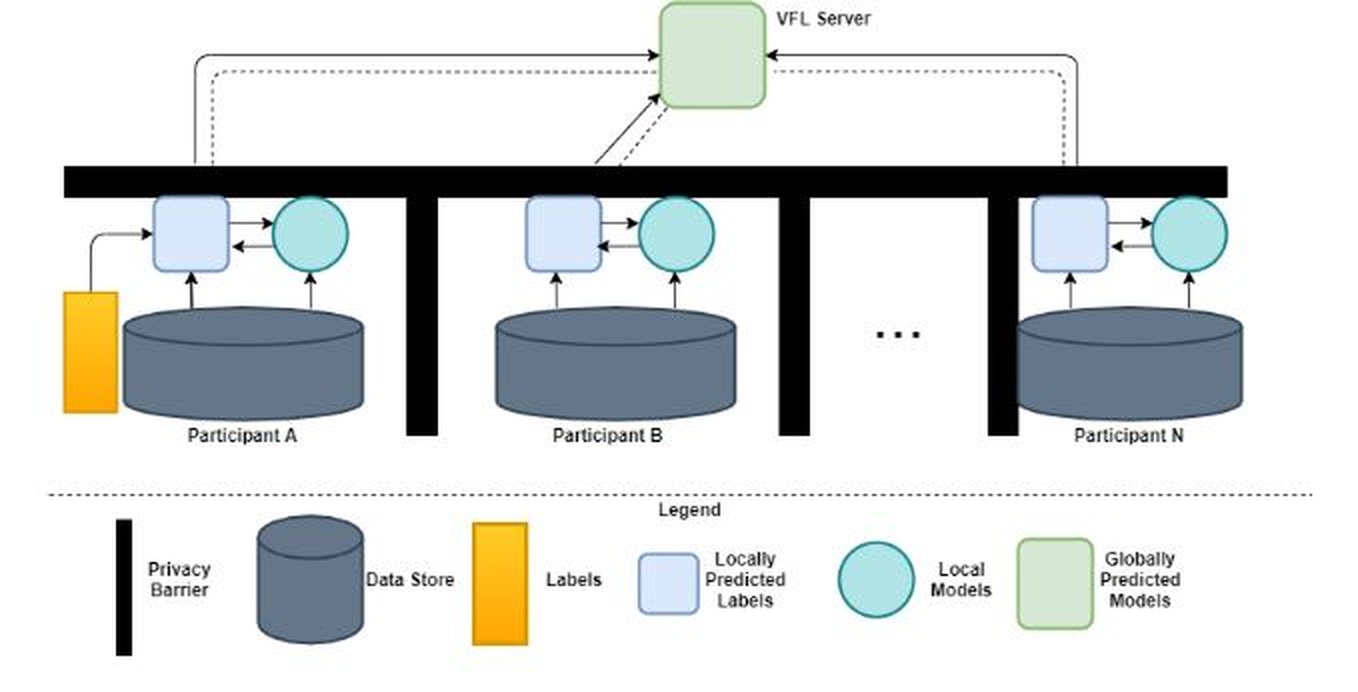}
\caption{The architecture of MMVFL \cite{aledhari2020federated}.}
\label{14}
\end{figure}

The second work by Siwei Feng and Han Yu proposes a new architecture based on the vertical FL system. The architecture proposed by the authors is specifically known as the Multi-Participant Multi-Class Vertical Federated Learning Framework (MMVFL). This particular architecture (Figure \ref{14}) is designed to manage multiple participants. The authors note that MMVFL enables the sharing of labels in a way that preserves the privacy of the owner and other participants. The assumption that records from different entities have the same feature space but may not be associated with the same sample ID space is problematic when introducing a horizontal fuzzy logic architecture. Unfortunately, this is not always the case, and the proposed structure aims to mitigate this drawback. The goal of the MMVFL framework is to learn a large number of frameworks to achieve different objectives. The goal is to increase the level of personalization in the learning process. The authors used two computer vision datasets to evaluate the performance of their framework: Additionally, the authors compare their framework with alternative approaches: the more features the framework includes, the better the results. They also observe that the MMVFL framework performs better the more features it uses \cite{feng2020multi}.

\begin{figure}
\centering
\includegraphics[height= 5 cm]{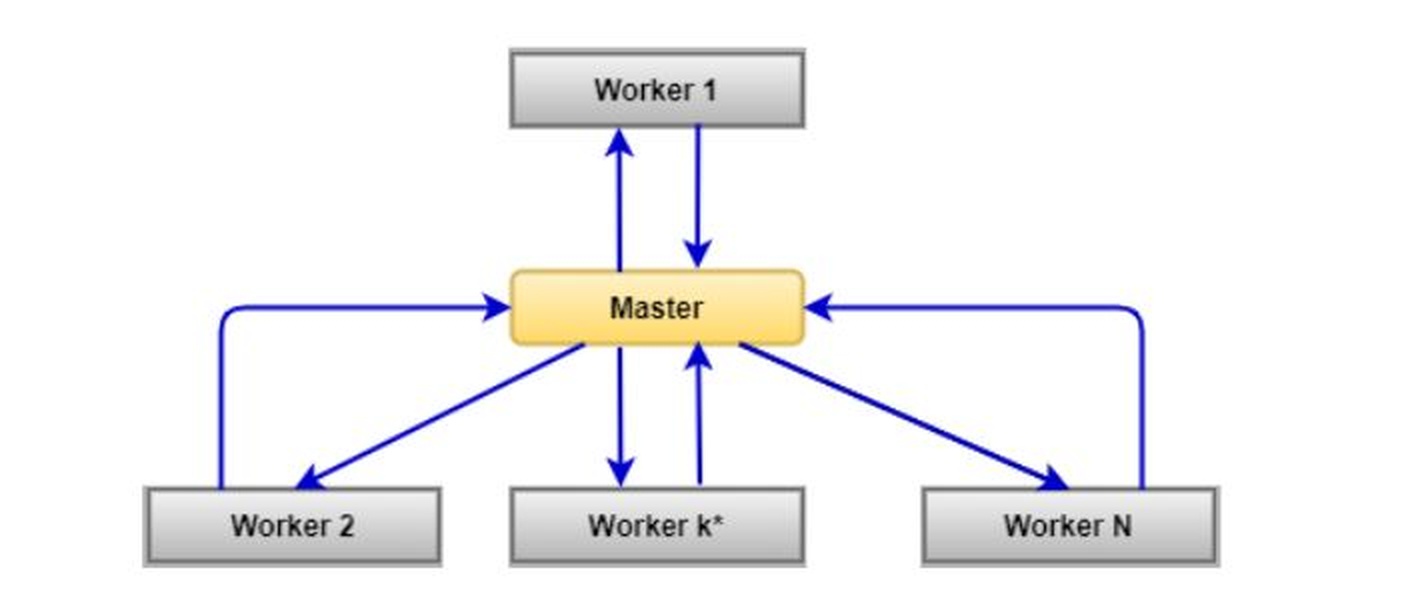}
\caption{The architecture of FEDF \cite{aledhari2020federated}.}
\label{15}
\end{figure}

FL's method is meant to allow for concurrent training while still protecting anonymity. A model may be trained on numerous geographically scattered training data sets—which may belong to different owners—using their framework, known as FEDF. As seen in Fig. \ref{15}, the authors' proposed design consists of a master and X workers. Additionally, the writers were able to test their framework on a variety of systems. The major datasets utilized to assess the FEDF architecture were the CIFAR-10 membrane data set (MEMBRANE) and the health care imaging data set (HEART-VESSEL). The assessment criteria were performance, training speed, and data volume transmitted.

\begin{figure}
\centering
\includegraphics[height= 5 cm]{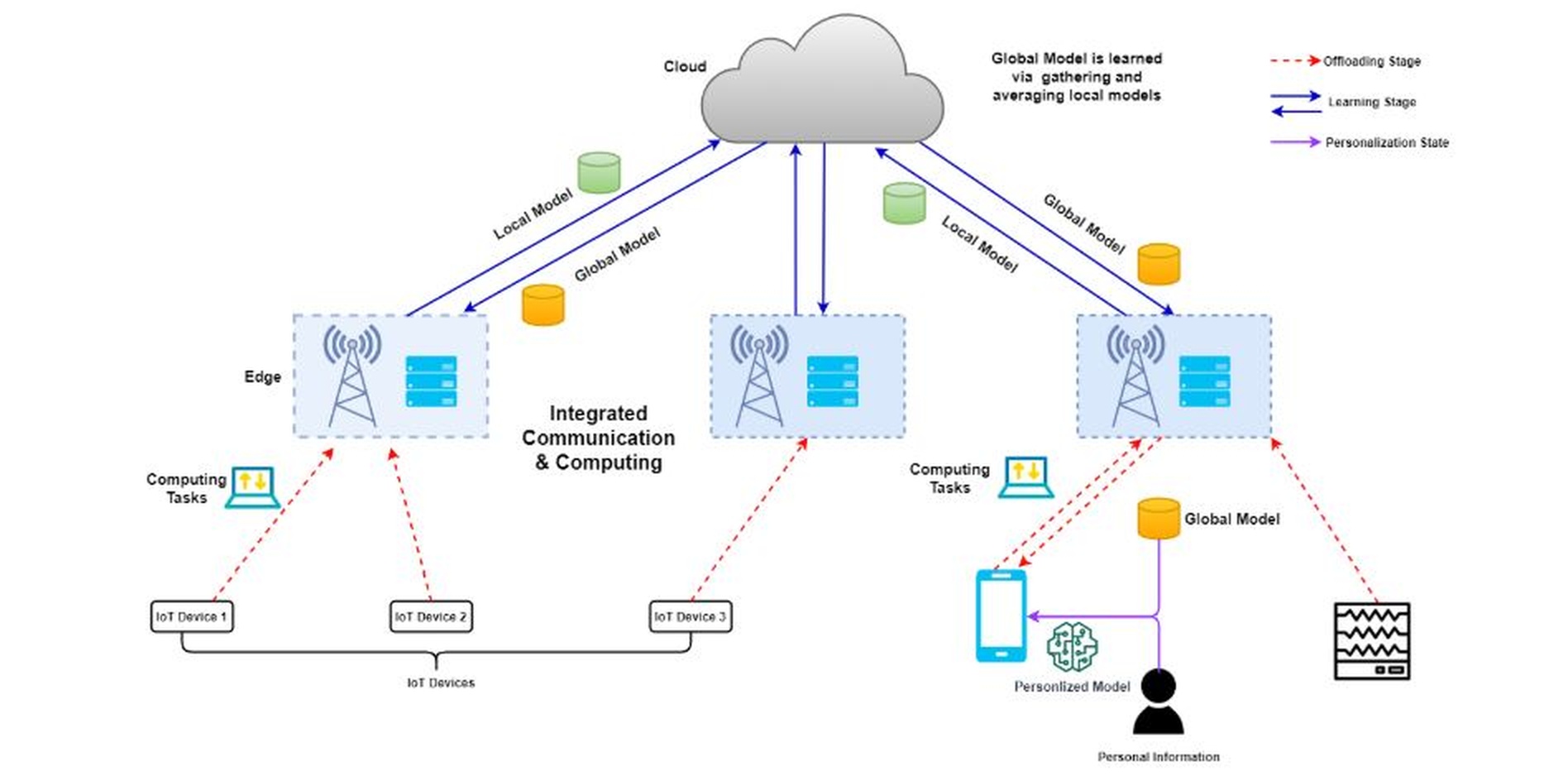}
\caption{The architecture of  PerFit \cite{aledhari2020federated}.}
\label{16}
\end{figure}

Figure \ref{16} shows the authors' PerFit framework. PerFit was designed to help with a few FL and IoT-related issues. Upon closer inspection, PerFit's cloud-based architecture should provide IoT devices with easily accessible processing capability, according to the authors. The architecture is set up in a way that allows any Internet of Things device to release its computational burden, thereby meeting the demands for low latency and efficiency.

The creation of a privacy-preserving FL application for anticipating customers' financial distress is the major focus of the study \cite{imteaj2022leveraging}. This approach addresses the resource and data privacy restrictions of traditional centralized machine learning models. The main contribution is a new FL method that allows partial task contributions, which reduces the effect of straggler agents and enhances model convergence and performance in resource-constrained contexts. The suggested approach outperforms current FL models in accuracy and maintains data privacy while achieving accuracy comparable to centralized models. The authors provided the general process flowchart in Figure \ref{17} and provided a step-by-step breakdown of their suggested methodology as follows:

\begin{figure}
\centering
\includegraphics[height= 12 cm]{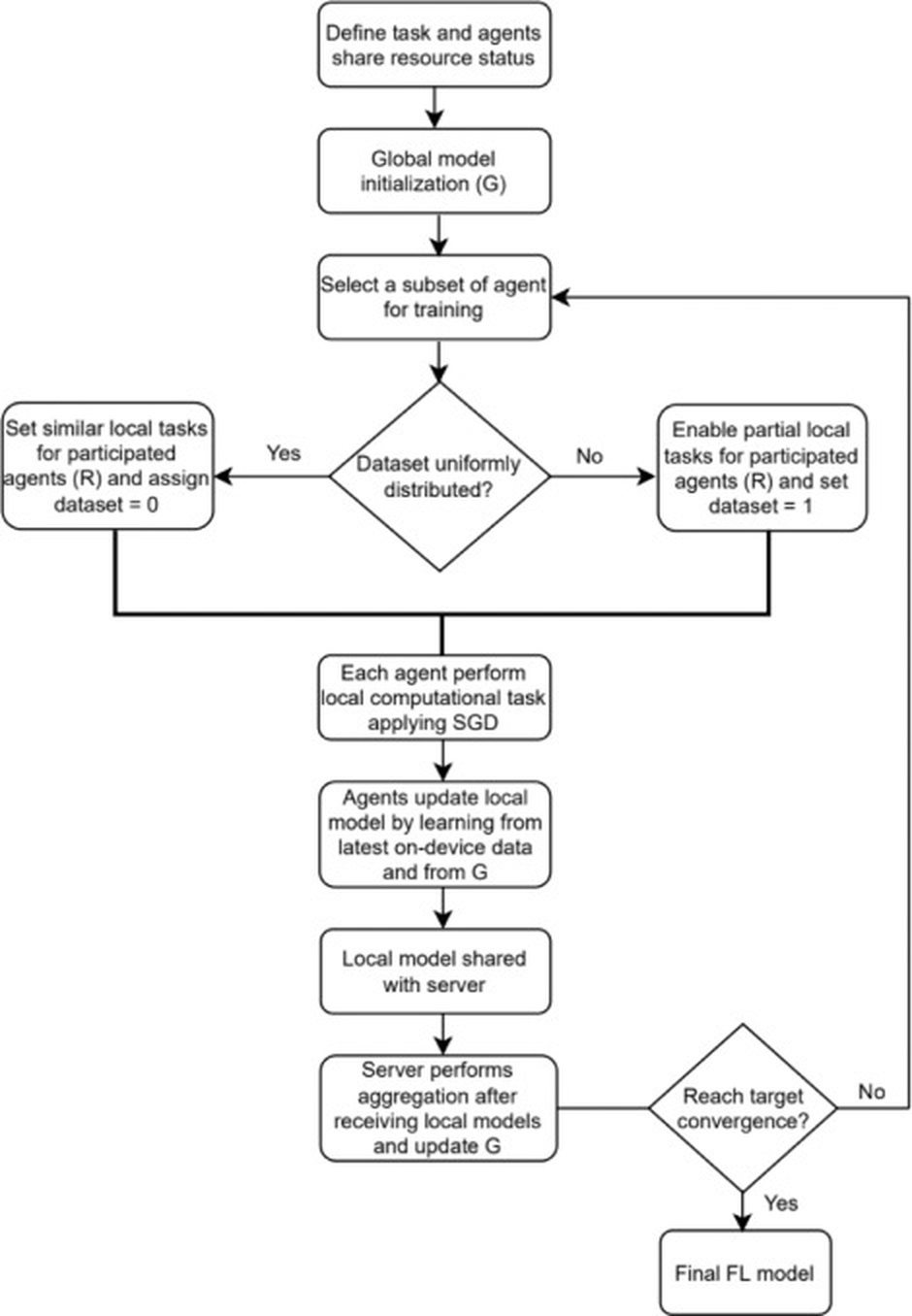}
\caption{Diagram illustrating the suggested FL approach for forecasting a customer's financial demise \cite{imteaj2022leveraging}.}
\label{17}
\end{figure}

\section{Federated Learning's Limitations and Difficulties}\label{sec2}
FL has several advantages, but its complete adoption across industries is hampered by a number of obstacles, particularly those related to privacy, security, and technical constraints. The fact that FL training data is inherently imperfect—it might be biased, uneven, or incomplete—is one of the main obstacles \cite{khan2021federated}. Poor model performance results from an uneven distribution of training samples among entities, a phenomenon known as data imbalance \cite{zhang2022federated}. The training process is made more difficult by missing classes, features, and values since distinct entities may have datasets that are missing crucial information, which leads to erroneous models. Moreover, the complexity is increased by the heterogeneity of data resulting from its dispersion across several places, rendering crude applications of FL models ineffectual.

Effective communication presents a big additional difficulty. FL uses a lot of devices, particularly in environments like healthcare \cite{singh2022framework}, \cite{ali2022federated} where privacy concerns make local data maintenance essential. In order to solve the slower communication speeds inherent in FL, it is imperative that the number of communication rounds and message sizes exchanged throughout the training process be reduced. Effective communication techniques are required to guarantee model updates in a timely manner without jeopardizing data privacy.

System heterogeneity adds still another level of complexity \cite{zhou2024distributed}. Stragglers—devices that are unable to keep up with the training process—can result from the varied processing capacities and network circumstances of participating devices, which delays the convergence of the model. Concerns about privacy are also very important since, whereas FL tries to keep sensitive data local, there is always a chance that information could leak during model upgrades \cite{makkar2023securefed}. To effectively apply FL across several industries, it is imperative to find creative solutions that improve data handling, communication efficiency, and privacy preservation.

\section{Future Research on FL and Zero Trust Security}
Combining Federated Learning (FL) with Zero Trust Architecture (ZTA) can greatly improve security, privacy, and efficiency in distributed AI systems. To ensure that no one party controls the training process completely, future research should look into using blockchain technology in FL. This would create secure, decentralized records for model updates, reducing the risk of harmful model changes and increasing trust and transparency.

Furthermore, adding differential privacy to FL-based ZTA can help protect sensitive information when model updates are shared. This approach is especially important in fields like healthcare, finance, and government, where data privacy and legal compliance are critical.

Additionally, real-time adaptive authentication using FL can improve security and access control by continuously checking users and devices based on their behavior patterns. By exploring these areas, FL-enabled ZTA can help create a safer, more resilient, and privacy-focused AI environment.

\section{Conclusion}\label{sec13}
FL transforms machine learning by decentralizing training among clients while ensuring data security and privacy. This article examines the core concepts of FL, evaluates various federated learning algorithms, and analyzes the architectures within the FL ecosystem. Our findings highlight FL's significant benefits, including enhanced privacy, regulatory compliance, and cost reductions, while also addressing challenges such as communication overhead, data heterogeneity, and privacy issues. The goal of this work is to advance FL research and applications by tackling these challenges and exploring future directions. We also present architectural principles to guide the development of effective federated learning systems, providing a resource for understanding the fundamental aspects of FL and paving the way for further advancements in this area.

\backmatter

\bmhead{Acknowledgements}

The authors would like to thank their friends and family for their continued support.





\bibliography{sn-bibliography}

\end{document}